\documentclass[11pt]{article}

\usepackage[final]{acl}

\usepackage{times}
\usepackage{latexsym}
\usepackage{xcolor}

\usepackage[T1]{fontenc}

\usepackage[utf8]{inputenc}

\usepackage{microtype}

\usepackage{inconsolata}

\usepackage{graphicx}
\usepackage{booktabs}
\usepackage{enumitem}
\usepackage{amsmath}

\usepackage{booktabs}
\usepackage{multirow}
\usepackage{amssymb}
\usepackage{pifont}
\usepackage[misc]{ifsym}
\usepackage{xcolor}
\newcommand{\cmark}{\ding{51}}
\newcommand{\xmark}{\ding{55}}

\title{ShredBench: Evaluating the Semantic Reasoning Capabilities of Multimodal LLMs in Document Reconstruction}

\author{
  \textbf{Zichun Guo\textsuperscript{\dag,1}, Yuling Shi\textsuperscript{\dag,2}, Wenhao Zeng\textsuperscript{2}, Chao Hu\textsuperscript{2},} \\
  \textbf{Haotian Lin\textsuperscript{2}, Terry Yue Zhuo\textsuperscript{3}, Jiawei Chen\textsuperscript{4}, Xiaodong Gu\textsuperscript{{\textrm{\Letter}},2}, Wenping Ma\textsuperscript{{\textrm{\Letter}},1}} \\
  \textsuperscript{1}Xidian University \quad \textsuperscript{2}Shanghai Jiao Tong University \\
  \textsuperscript{3}Alibaba Qwen \quad \textsuperscript{4}Old Dominion University \\
  \texttt{guozichun3@gmail.com, xiaodong.gu@sjtu.edu.cn, wp\_ma@mail.xidian.edu.cn}
}

\begin{document}
\maketitle
{\renewcommand{\thefootnote}{}
\footnotetext{\textsuperscript{\dag}Equal contribution. {\textrm{\Letter}}\,Corresponding author.}}

\begin{abstract}
Multimodal Large Language Models (MLLMs) have achieved remarkable performance in Visually Rich Document Understanding (VRDU) tasks, but their capabilities are mainly evaluated on pristine, well-structured document images. We consider content restoration from shredded fragments, a challenging VRDU setting that requires integrating visual pattern recognition with semantic reasoning under significant content discontinuities. To facilitate systematic evaluation of complex VRDU tasks, we introduce \textsc{ShredBench}, a benchmark supported by an automated generation pipeline that renders fragmented documents directly from Markdown. The proposed pipeline ensures evaluation validity by allowing the flexible integration of latest or unseen textual sources to prevent training data contamination. \textsc{ShredBench} assesses four scenarios (English, Chinese, Code, Table) with three fragmentation granularities (8, 12, 16 pieces). Empirical evaluations on state-of-the-art MLLMs reveal a significant performance gap: The method is effective on intact documents; however, once the document is shredded, restoration becomes a significant challenge, with NED dropping sharply as fragmentation increases. Our findings highlight that current MLLMs lack the fine-grained cross-modal reasoning required to bridge visual discontinuities, identifying a critical gap in robust VRDU research\footnote{Code and dataset are available at \url{https://github.com/ythere-y/ShredBench}.}.

\end{abstract}

\section{Introduction}

\begin{figure}[t!]
    \centering
    \includegraphics[width=\linewidth]{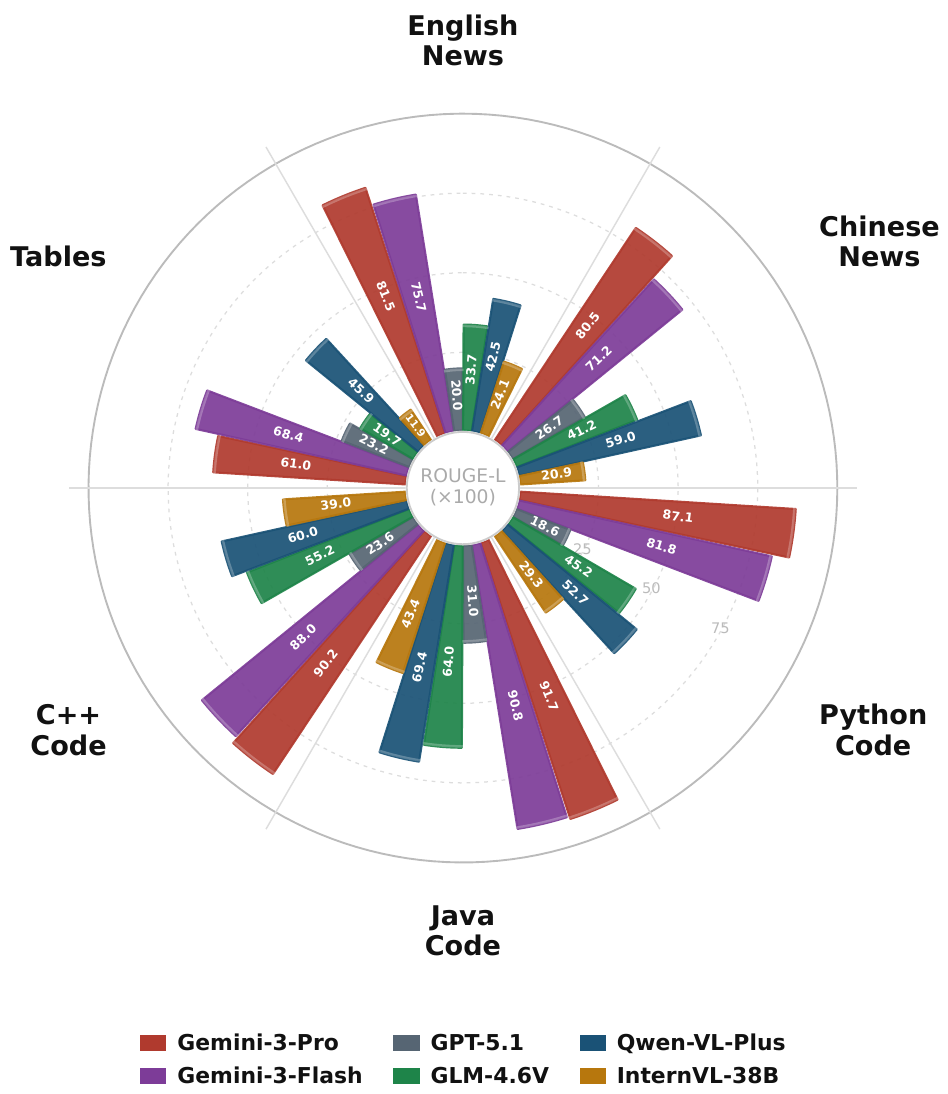}
    \caption{Evaluation results on \textsc{ShredBench} across 6 dimensions (Metric: ROUGE-L). Our proposed benchmark reveals significant gaps in current MLLMs' capabilities on fragmented documents.} 
    \label{fig:radar_chart}
\end{figure}

The advance in Multimodal Large Language Models (MLLMs), such as GPT-5~\cite{openai2025gpt5} and Gemini 3 Pro ~\cite{google2025gemini}, has revolutionized the field of Visually Rich Document Understanding (VRDU) ~\cite{yin2024survey,wang2023vrdu,wang2025vrag,wang2025vidorag}. By projecting visual features into a shared semantic space with textual representations, these models have almost achieved human expert performance on tasks ranging from standard Optical Character Recognition (OCR)~\cite{lee2023pix2struct, lv2023kosmos} to complex information extraction (CIE) from well-formatted documents ~\cite{kim2022ocr, yu2023structextv2, tang2023unifying}. However, real-world document processing often encounters inputs that are far from ideal, where documents may be occluded, damaged, or physically torn. Although recent high-resolution MLLMs ~\cite{wang2023cogvlm, li2023monkey, chen2025autoneuralcodesigningvisionlanguagemodels, chen2025progressive} attempt to mitigate visual noise and enhance fine-grained perception, the specific challenge of reconstructing physically fragmented information remains underexplored. While recent benchmarks have begun to address robustness against image corruptions ~\cite{qiu2025benchmarking} or super-long context retrieval ~\cite{chia2024mlongdoc}, the challenge of reconstructing physically fragmented information remains underexplored. While humans can rely on strong language priors and world knowledge ~\cite{wagemans2012century, schlichting2015memory} to mentally piece together fragmented information, the extent to which MLLMs possess this capability remains an open question.

In this paper, we explore \emph{shredded content restoration} at the intersection of vision and NLP. Unlike traditional jigsaw puzzles based on edge matching, this task demands profound semantic reasoning~\cite{wang2024readandthink}. For instance, connecting \textit{``The algorithm optimiz-''} with \textit{``-es the loss function''} relies less on ambiguous visual cuts than on syntactic expectation. Consequently, this task serves as a rigorous probe for evaluating whether MLLMs can leverage internal language priors to maintain coherence across visual discontinuities.

To systematically evaluate this, we propose \textsc{ShredBench}, a benchmark characterized by three key dimensions:
\textit{(1) Multi-Granularity Complexity.} We partition images into 8, 12, and 16 fragments. This hierarchy enables the analysis of how visual entropy correlates with performance degradation.
\textit{(2) Diverse Scenarios.} Comprising 756 documents, our dataset spans English and Chinese text, source code (strict syntax), and tables (complex 2D structure). Tables and code are notably difficult, requiring models to restore rigid indentation and alignment—a challenge even for specialized models~\cite{zhang2024tablellama}.
\textit{(3) Extensive Experiments.} We evaluate state-of-the-art proprietary and open-source MLLMs. Using standard textual metrics, we establish the first quantitative baselines to facilitate future research.

We employ NED, TEDS, BLEU, and ROUGE-L as our primary evaluation metrics and conduct extensive experiments across 14 representative MLLMs, including both leading proprietary and open-source models. The results are sobering: While models exhibit high proficiency on intact documents, their performance collapses under fragmentation. In the hardest setting (16 fragments), the average NED reaches a high of 0.73, even the most advanced models failing to identify correct reading orders or hallucinating non-existent bridging text~\cite{guan2023hallusionbench, li2023evaluating}. Our study reveals that current MLLMs struggle to effectively align visual positional embeddings with semantic continuity, often treating fragments as independent entities rather than parts of a cohesive whole. 

Our contributions are summarized as follows. First, we introduce \textsc{ShredBench}, the first benchmark specifically designed to stress-test the semantic reasoning capabilities of MLLMs via shredded content restoration. Second, we design an automated pipeline for generating shredded document benchmarks with adjustable granularity. This enables the synthesis of diverse samples covering English and Chinese text, source code, and tables, thereby presenting a comprehensive range of semantic and structural challenges. Third, we conduct a comprehensive evaluation of various MLLMs, revealing significant limitations in their ability to handle visual structural noise and maintain coherence in both textual semantics and 2D spatial layouts.

\section{Related Work}
\label{sec:related_work}

\subsection{Benchmarking Multimodal Reasoning}
\label{sec:benchmarks}

Recent MLLM benchmarks have expanded beyond visual perception to evaluate complex reasoning~\cite{hu2025beyondemotion,dai2026cedar,li2026wrote}. Representative works include MMBench~\cite{liu2023mmbench} and SEED-Bench~\cite{li2023seed} for general and generative comprehension, alongside domain-specific benchmarks like MathVista~\cite{lu2024mathvista} that target mathematical and logical deduction.
However, these benchmarks largely focus on coherent and clean inputs, leaving models' ability to reason under structurally disordered or fragmented data underexplored. 
In contrast, \textsc{ShredBench} is specifically designed to evaluate semantic reconstruction in the presence of structural disruption, providing a rigorous assessment of long-context coherence under disordered inputs.

\begin{table*}[t]
\centering
\label{tab:benchmark_comparison}  
\resizebox{\textwidth}{!}{%
\begin{tabular}{l|c|c|c|c|c|cc}
\toprule
\multirow{2}{*}{\textbf{Benchmark}} & \multirow{2}{*}{\textbf{Domain}} & \multirow{2}{*}{\textbf{Modality}} & \multirow{2}{*}{\textbf{Deformation}} & \multirow{2}{*}{\textbf{Reasoning Type}} & \multirow{2}{*}{\textbf{Granularity}} & \multicolumn{2}{c}{\textbf{Capabilities}} \\
 & & & & & & OCR & Reconst. \\
\midrule
\multicolumn{8}{l}{\textit{Document Parsing Benchmarks}} \\
\midrule
\textbf{OmniDocBench} ~\cite{ouyang2025omnidocbench} & Document & Text, Table, Formula & / & Structural Parsing & /  & \cmark & \xmark  \\
\textbf{WildDoc} ~\cite{wang2025wilddoc} & Scene Doc & Text, Chart & Shadow, Blur, Warp & Robust Perception & / & \cmark & \xmark \\
\textbf{DocPTBench} ~\cite{du2025docptbench} & Photo Doc & Text & Geom. Warp & Parsing \& Trans.  & / & \cmark & \xmark \\
\midrule
\multicolumn{8}{l}{\textit{Visual Jigsaw \& Reconstruction Benchmarks}} \\
\midrule
\textbf{Jigsaw-Puzzles} ~\cite{lyu2025jigsaw} & Natural Img & Visual Pixels & Grid Crop (2D) & Spatial Arrangement & Grid (2x2 to 5x5) & \xmark & \cmark \\
\textbf{RePAIR} ~\cite{tsesmelis2024repair} & Artifacts & 3D Geometry & Erosion, Fragments & Geometric Matching & / & \xmark & \cmark \\
\midrule
\multicolumn{8}{l}{\textit{Proposed Benchmark}} \\
\midrule
\textbf{ShredBench (Ours)} & Hybrid & Text, Table, Code & 3D Shredding & Semantic Bridging & Voronoi (8, 12, 16 pcs) & \cmark & \cmark \\
\bottomrule
\end{tabular}%
}
\caption{Comparison of ShredBench with representative benchmarks. Domain: target data domain. Modality: input data types. Deformation: visual or physical distortion applied to inputs. Reasoning Type: core cognitive ability evaluated. Granularity: fragment or subunit size/layout. Capabilities: evaluated capabilities, including OCR and implicit reconstruction reasoning.}
\end{table*}
\vspace{-0.2cm}

\subsection{Document Parsing and Understanding}
\label{sec:doc_parse_understand}

The field has evolved from modular OCR to end-to-end MLLMs capable of holistic parsing and understanding. In \textit{document parsing}, models like Nougat~\cite{blecher2023nougat} reconstruct papers into markup, while TextMonkey~\cite{liu2024textmonkey} and Vary~\cite{wei2023vary} handle dense text and layout reconstruction. For \textit{document understanding}, proprietary models such as GPT-5~\cite{openai2025gpt5} and Gemini 3 Pro~\cite{google2025gemini} show strong zero-shot reasoning, while open-source models like LLaVA~\cite{liu2023llava}, Qwen-VL~\cite{bai2023qwenvl}, and InternVL~\cite{chen2024internvl} focus on high-resolution processing and reducing hallucinations.

Comprehensive benchmarks support these tasks: OmniDoc~\cite{ouyang2025omnidocbench} and HierText~\cite{long2022towards} target multi-task reconstruction and dense text perception, DocVQA~\cite{mathew2021docvqa} and ChartQA~\cite{masry2022chartqa,cheng2026enhancingfinancialreportquestionanswering} assess information extraction and logical reasoning, and WildDoc~\cite{wang2025wilddoc} evaluates MLLMs on natural scene documents with lighting and physical distortions, revealing robustness limitations.

However, these approaches predominantly assume clear, intact inputs, ignoring scenarios where document structure is physically disrupted. Consequently, the ability of MLLMs to reason over fragmented or shredded documents remains underexplored. \textsc{ShredBench} addresses this gap by evaluating semantic reconstruction under structural disruption, advancing research into physically impaired document understanding.

\subsection{Visual Reconstruction}
\label{sec:visual_recon}

Visual reconstruction has traditionally been framed as the \textit{Jigsaw Puzzle} problem in computer vision. In the image domain, traditional methods use edge detection or Deep Metric Learning~\cite{noroozi2016unsupervised, paixao2020faster}, and neural approaches like PairingNet~\cite{zhou2023pairingnet} leverage graph networks and transformers for improved matching. Benchmarks such as Jigsaw-Puzzles~\cite{lyu2025jigsaw} and RePAIR~\cite{tsesmelis2024repair} assess spatial reasoning on natural images and fragmented artifacts, but focus primarily on visual or geometric cues.

However, shredded content restoration adds challenges due to sparse text and uniform backgrounds, where visual cues are ambiguous. Semantic reasoning—completing truncated text or formulas—is essential. \textsc{ShredBench} evaluates this capability, testing scenarios beyond the reach of purely visual methods.

\section{ShredBench Dataset}

\begin{figure*}[t]
    \centering
    \includegraphics[width=0.9\textwidth]{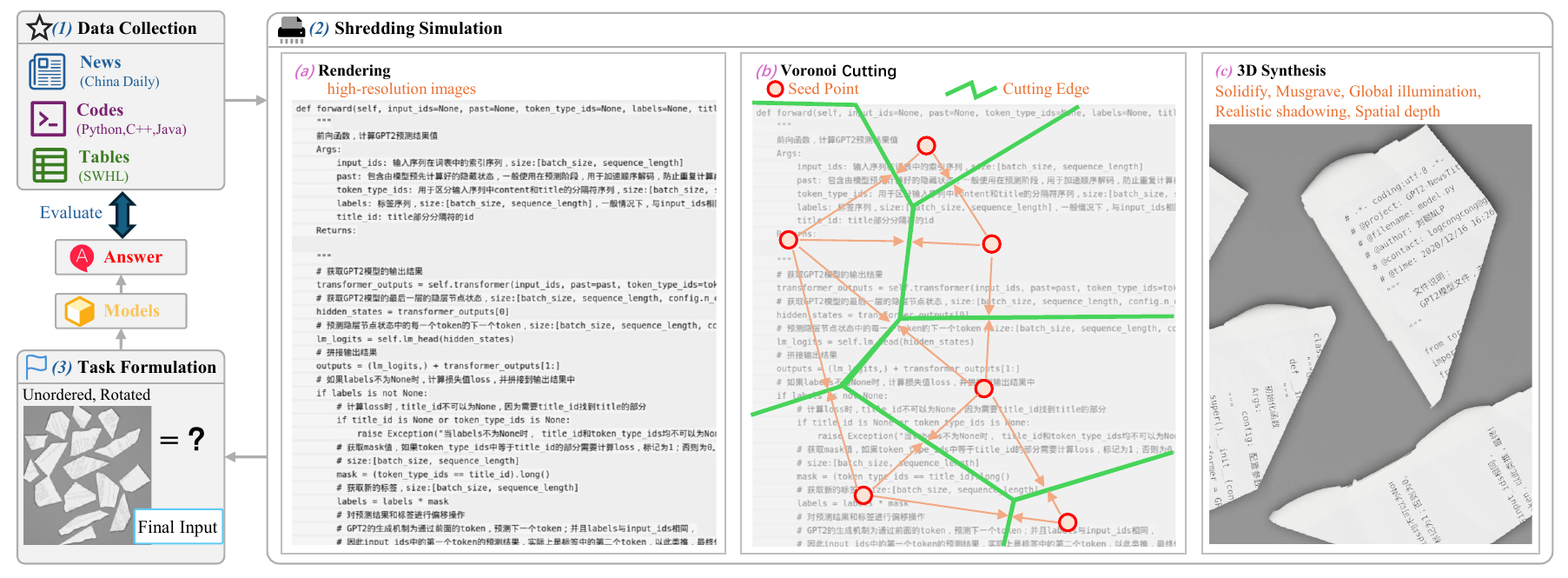}
    \caption{Schematic illustration of the \textsc{ShredBench} data generation pipeline. The process consists of three stages: (1) Data Collection from diverse sources (News, Code, Tables), (2) Shredding Simulation including Voronoi tessellation and physics-based 3D rendering, and (3) Task Formulation where the unordered fragments serve as the final input.}
    \label{fig:bad_case_study}
\end{figure*}

In this section, we present the construction process of \textsc{ShredBench}. Our pipeline consists of three stages: content acquisition across multiple domains, physics-based shredding simulation, and the formulation of the reconstruction task.

\subsection{Data Collection} 
Data leakage is a widely discussed concern in benchmark evaluation~\cite{fang-etal-2025-lastingbench,chen2026schr}. To ensure the model's robustness across different semantic contexts and layouts, we constructed a diverse corpus comprising bilingual news, programming code, and scientific tables.

\paragraph{News Articles.} We collected high-quality journalism text to represent standard natural language prose. 
For English content, we scraped articles from \textit{China Daily} via RSS feeds (covering World, Business, and Opinion sections). 
For Chinese content, we sourced articles from \textit{People.com.cn} (People's Daily Online). 
To ensure content density, we filtered articles with lengths between 800 and 2,500 characters.

\paragraph{Source Code.} Code has emerged as a key evaluation domain for LLMs across diverse tasks including generation, understanding, compression, and others~\cite{shi2024code,shi2024between,peng2025swe,hu2026line,wang2026effiskill,zeng2025pruning,shi2025longcodezip,Ma_2026,shi2026code,pmlr-v306-xie26j,chen2026classeval,shi2026swe,wang-etal-2026-context}. To introduce structured syntax and indentation challenges, we utilized the GitHub API to crawl code snippets in three major programming languages: \texttt{Python}, \texttt{C++}, and \texttt{Java}. We specifically targeted files with sizes between 1KB and 4KB and extracted metadata (e.g., commit dates) to enrich the dataset context.

\paragraph{Scientific Tables.} To introduce structured data challenges, we sourced tabular samples from the public SWHL table recognition dataset\footnote{\url{https://huggingface.co/datasets/SWHL/table_rec_test_dataset}}. This dataset aggregates a diverse range of table layouts, including bordered and borderless styles, complex headers, and spanning cells. Incorporating these samples ensures that \textsc{ShredBench} rigorously evaluates the model's capacity to reconstruct strict spatial dependencies and grid-like structures typical in academic and financial documents.

\subsection{Shredding Simulation} 
Standard 2D cropping preserves pixel-perfect con-
tinuity, allowing models to bypass semantic rea-
soning by exploiting trivial edge matches. To rig-
orously benchmark document understanding, we
developed a physics-based rendering pipeline that
simulates real-world artifacts, including crumpling,
shadows, and irregular edges. This approach sup-
presses visual shortcuts, ensuring that successful
reconstruction depends on interpreting the seman-
tic context.

\begin{figure}[h] 
    \centering
    \includegraphics[width=1.0\linewidth]{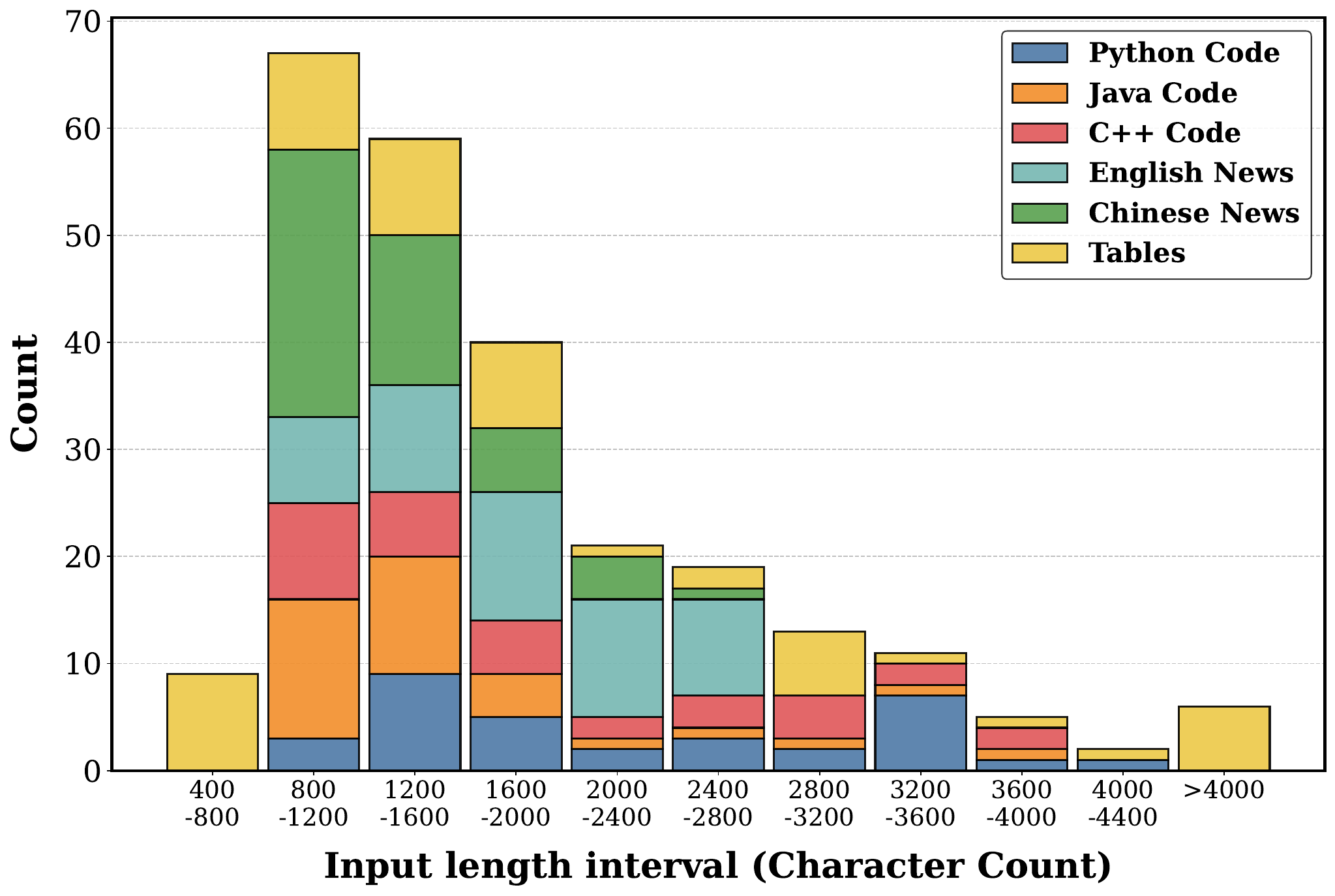}
    
    \caption{Distribution of dataset input lengths (in characters). The dataset is segmented into intervals of 400 characters, showing the count of files for each category (Code, News, Tables).}
    
    \label{fig:length_dist}
\end{figure}
\vspace{-0.2cm}

\paragraph{Document Rendering.} First, raw text data is rendered into high-resolution images ($1600$px width) using a headless Chrome browser. We apply custom CSS styling (Times New Roman/SimSun fonts, 28px size) and inject random RGB noise to simulate paper texture.

\paragraph{Voronoi Cutting Algorithm.} To generate realistic, irregular fragments, we employ a Voronoi tessellation approach. For a given document image, we randomly sample $N$ seed points ($N \in \{8, 12, 16\}$) on the canvas. A $k$-d tree algorithm assigns each pixel to the nearest seed point, naturally forming jagged, non-rectilinear boundaries that mimic manual shredding.

\paragraph{3D Physical Synthesis.} The 2D fragments are then imported into Blender for physical simulation. We apply a \textit{Solidify} modifier (thickness $0.002$) and distinct displacement maps: a \textit{Marble} texture for large-scale waves and a \textit{Musgrave} texture for sharp crumples. The fragments are scattered using a pixel-perfect packing algorithm to ensure no overlap. Finally, the scene is rendered using the Cycles engine at 4K resolution ($4096 \times 4096$) with global illumination, creating realistic shadowing and spatial depth.

\subsection{Quality Control}
To ensure the rigorousness of \textsc{ShredBench}, we implemented a verification process on a random sample of 50 documents. Two independent human annotators assessed whether the fragments contained sufficient semantic cues for unique reconstruction. The inspection yielded a Cohen's Kappa ($\kappa$)\cite{cohen1960coefficient} of 0.79, indicating substantial inter-annotator agreement and confirming the objective nature of the task. Crucially, final adjudication confirmed that 96\% of the sampled fragments (48/50) were strictly solvable, while only a marginal fraction (4\%) was deemed ambiguous and subsequently removed. Although a minor noise floor exists, it is statistically negligible compared to the drastic performance collapse observed in state-of-the-art MLLMs (avg. NED 0.73), confirming that the reported failure stems from model reasoning limitations rather than data defects.
\vspace{-0.3cm}

\subsection{Task Formulation}
We formulate the shredded content restoration problem as a set-to-sequence task.
Formally, let $\mathcal{I} = \{f_1, f_2, \dots, f_N\}$ be a set of unordered, scattered image fragments derived from a single source document $D$. 
The input to the model is the visual set $\mathcal{I}$, where each fragment $f_i$ contains partial visual information, potentially rotated and subjected to lighting distortions.

The objective is to generate a text string $\hat{T}$ that matches the ground-truth text content $T$ of the original document $D$. 
Unlike geometric reconstruction tasks that require predicting the spatial coordinates $(x, y, \theta)$ of each piece, our task focuses purely on content restoration. The model must implicitly solve the jigsaw puzzle to recover the correct reading order and utilize OCR capabilities to transcribe the text.

\section{Experimental Setup} 
\label{sec:setup}

\subsection{Models Evaluated}
To ensure a comprehensive evaluation across different architectures and capabilities, we selected a diverse set of MLLMs, ranging from proprietary state-of-the-art model APIs to leading open-source model weights. 
\paragraph{Proprietary Models:}
We select GPT-5 Mini and GPT-5.1~\cite{openai2025gpt5} as representative baselines for efficiency and high-level reasoning, respectively. Similarly, we evaluate Google's Gemini 3 Flash for low-latency tasks and Gemini 3 Pro~\cite{google2025gemini} for state-of-the-art multimodal logic. 

\paragraph{Open-Source Models:}
InternVL~~\cite{chen2024internvl} and Qwen-VL series (Plus/Flash)~~\cite{bai2023qwenvl} serve as robust general-purpose baselines with strong visual understanding. 
For specialized capabilities, we include GLM-4.6v~~\cite{glm2024glm4} for bilingual interactions, and Mistral3-Reasoning~~\cite{mistral2025magistral} for transparent multi-step logic. 
In the domain of document parsing, we evaluate DeepSeek-OCR~~\cite{wu2024deepseekvl2}, which utilizes an MoE visual encoder for high-resolution processing, and Hunyuan-OCR~~\cite{tencent2025hunyuanocr}, optimized for end-to-end text spotting.

\begin{table*}[t!]
    \centering
    \tiny 
    \setlength{\tabcolsep}{3.5pt} 
    \renewcommand{\arraystretch}{0.9}
    \resizebox{0.8\textwidth}{!}{%
    \begin{tabular}{l|ccc|ccc|ccc}
        \toprule
        \multirow{2}{*}{\textbf{Model}} 
        & \multicolumn{3}{c|}{\textbf{8 Fragments}} 
        & \multicolumn{3}{c|}{\textbf{12 Fragments}} 
        & \multicolumn{3}{c}{\textbf{16 Fragments}} \\
        \cmidrule(lr){2-4} \cmidrule(lr){5-7} \cmidrule(lr){8-10}
        & NED$\downarrow$ & BLEU$\uparrow$ & ROUGE$\uparrow$ 
        & NED$\downarrow$ & BLEU$\uparrow$ & ROUGE$\uparrow$ 
        & NED$\downarrow$ & BLEU$\uparrow$ & ROUGE$\uparrow$ \\
        \midrule
        \multicolumn{10}{c}{\textit{\textbf{Open-source Models}}} \\
        \midrule
        InternVL3.5-8B
        & 0.78 & 0.07 & 0.24 & 0.79 & 0.05 & 0.21 & 0.78 & 0.05 & 0.21 \\
        
        InternVL3.5-14B
        & 0.76 & 0.08 & 0.26 & 0.77 & 0.07 & 0.24 & 0.78 & 0.07 & 0.23 \\
        
        InternVL3.5-38B
        & 0.74 & 0.10 & 0.28 & 0.75 & 0.08 & 0.26 & 0.76 & 0.08 & 0.24 \\

        Mistral3-Reas-8B
        & 0.77 & 0.09 & 0.28 & 0.79 & 0.06 & 0.23 & 0.79 & 0.06 & 0.24 \\

        Mistral3-Reas-14B
        & 0.76 & 0.10 & 0.30 & 0.77 & 0.09 & 0.28 & 0.77 & 0.08 & 0.27 \\

        DeepSeek-OCR
        & 0.86 & 0.02 & 0.12 & 0.87 & 0.01 & 0.09 & 0.87 & 0.01 & 0.10 \\

        Hunyuan-OCR
        & 0.88 & 0.01 & 0.15 & 0.88 & 0.01 & 0.14 & 0.89 & 0.00 & 0.12 \\

        GLM-4.6v
        & 0.67 & 0.20 & 0.45 & 0.70 & 0.17 & 0.40 & 0.71 & 0.15 & 0.37 \\
        
        Qwen-VL-Flash
        & 0.59 & 0.26 & 0.58 & 0.63 & 0.22 & 0.54 & 0.65 & 0.19 & 0.50 \\
        
        Qwen-VL-Plus
        & 0.59 & 0.26 & 0.58 & 0.63 & 0.22 & 0.53 & 0.65 & 0.20 & 0.50 \\

        \midrule
        \multicolumn{10}{c}{\textit{\textbf{Proprietary Models}}} \\
        \midrule
        GPT-5 Mini      
        & 0.86 & 0.06 & 0.27 & 0.84 & 0.04 & 0.26 & 0.84 & 0.05 & 0.25 \\
        
        GPT-5.1         
        & 0.77 & 0.07 & 0.28 & 0.81 & 0.05 & 0.22 & 0.82 & 0.04 & 0.21 \\
        
        Gemini 3 Flash  
        & 0.34 & 0.47 & 0.82 & 0.40 & 0.44 & 0.77 & 0.44 & 0.41 & 0.74 \\
        
        Gemini 3 Pro 
        & \textbf{0.33} & \textbf{0.51} & \textbf{0.83} 
        & \textbf{0.37} & \textbf{0.48} & \textbf{0.81} 
        & \textbf{0.41} & \textbf{0.44} & \textbf{0.76} \\
        \bottomrule
    \end{tabular}%
    }
    \caption{Overall Performance Summary. Aggregated results across all categories. The metrics are split into separate columns for clarity: NED ($\downarrow$), BLEU ($\uparrow$), and ROUGE ($\uparrow$). Gemini 3 Pro shows consistent superiority across all settings.}
    \label{tab:summary_results_split}
\end{table*}

\subsection{Evaluation Metrics} 
We employ three standard metrics to quantitatively evaluate the similarity between the generated text and the ground truth. Let $Y$ denote the ground truth (reference) text and $\hat{Y}$ denote the generated text (hypothesis).

\paragraph{NED and TEDS:}
We employ Normalized Edit Distance (NED)~\cite{levenshtein1965binary} for general text similarity. It normalizes the Levenshtein distance ($Lev$) between prediction $\hat{Y}$ and ground truth $Y$:
\begin{equation}
    NED(Y, \hat{Y}) = \frac{Lev(Y, \hat{Y})}{\max(|Y|, |\hat{Y}|)}
\end{equation}
A lower NED implies higher similarity. For tables, we use Tree-Edit-Distance-based Similarity (TEDS)~\cite{zhong2020image}, which models content as trees (e.g., HTML DOM) to assess both structure and accuracy:
\begin{equation}
    TEDS(T, \hat{T}) = 1 - \frac{TED(T, \hat{T})}{\max(|T|, |\hat{T}|)}
\end{equation}
where $TED(\cdot)$ is the tree edit distance; higher scores indicate better reconstruction.

\paragraph{BLEU (Bilingual Evaluation Understudy):} 
Proposed by~\citet{papineni2002bleu}, BLEU calculates the geometric mean of n-gram precision, penalized for brevity:
    \begin{equation}
        BLEU = BP \cdot \exp\left( \sum_{n=1}^N w_n \log p_n \right)
    \end{equation}
where $p_n$ is n-gram precision and $w_n$ are weights. The Brevity Penalty ($BP$) accounts for generation length bias:
    \begin{equation}
        BP = \begin{cases} 
        1 & \text{if } c > r, \\
        e^{(1 - r/c)} & \text{if } c \leq r, 
        \end{cases}
    \end{equation}
with $c$ and $r$ denoting generated and reference lengths, respectively.

\paragraph{ROUGE-L:} 
We use ROUGE-L~\cite{lin2004rouge} to capture sentence-level structure via the Longest Common Subsequence (LCS). Precision ($P_{lcs}$) and recall ($R_{lcs}$) are defined as:
    \begin{equation}
        R_{lcs} = \frac{LCS(Y, \hat{Y})}{|Y|}, \quad P_{lcs} = \frac{LCS(Y, \hat{Y})}{|\hat{Y}|}
    \end{equation}
The final score is the weighted F-measure of these components:
\begin{equation}
    ROUGE-L = \frac{(1 + \beta^2) R_{lcs} P_{lcs}}{R_{lcs} + \beta^2 P_{lcs}}
\end{equation}
where $\beta$ controls the relative importance of precision versus recall.

\begin{table*}[h!]
    \centering
    \scriptsize
    \resizebox{0.8\textwidth}{!}{%
    \begin{tabular}{l|ccc|ccc}
        \toprule
        \textbf{Model} 
        & \multicolumn{3}{c|}{\textbf{English News}} 
        & \multicolumn{3}{c}{\textbf{Chinese News}} \\
        & $N=8$ & $N=12$ & $N=16$ & $N=8$ & $N=12$ & $N=16$ \\
        \midrule
        \multicolumn{7}{c}{\textit{\textbf{Open-source Models}}} \\
        \midrule
        InternVL3.5-8B
        & 0.75 / 0.07 / 0.18 & 0.77 / 0.06 / 0.18 & 0.77 / 0.04 / 0.15 
        & 0.91 / 0.01 / 0.30 & 0.92 / 0.02 / 0.22 & 0.91 / 0.02 / 0.21 \\
        
        InternVL3.5-14B
        & 0.73 / 0.11 / 0.22 & 0.73 / 0.11 / 0.25 & 0.75 / 0.06 / 0.19 
        & 0.92 / 0.01 / 0.26 & 0.92 / 0.01 / 0.20 & 0.93 / 0.01 / 0.22 \\
        
        InternVL3.5-38B
        & 0.70 / 0.14 / 0.27 & 0.71 / 0.13 / 0.25 & 0.74 / 0.07 / 0.20 
        & 0.92 / 0.01 / 0.24 & 0.92 / 0.01 / 0.20 & 0.92 / 0.01 / 0.19 \\

        Mistral3-Reas-8B
        & 0.70 / 0.14 / 0.30 & 0.71 / 0.11 / 0.25 & 0.72 / 0.08 / 0.22 
        & 0.94 / 0.04 / 0.23 & 0.95 / 0.02 / 0.16 & 0.96 / 0.02 / 0.17 \\

        Mistral3-Reas-14B
        & 0.69 / 0.17 / 0.29 & 0.71 / 0.16 / 0.28 & 0.71 / 0.13 / 0.25 
        & 0.93 / 0.05 / 0.26 & 0.94 / 0.03 / 0.24 & 0.94 / 0.03 / 0.25 \\

        DeepSeek-OCR
        & 0.80 / 0.03 / 0.13 & 0.82 / 0.02 / 0.12 & 0.83 / 0.01 / 0.10 
        & 0.95 / 0.00 / 0.13 & 0.95 / 0.01 / 0.10 & 0.95 / 0.01 / 0.13 \\

        Hunyuan-OCR
        & 0.88 / 0.02 / 0.08 & 0.85 / 0.02 / 0.08 & 0.88 / 0.01 / 0.07 
        & 0.92 / 0.01 / 0.27 & 0.94 / 0.01 / 0.23 & 0.94 / 0.00 / 0.24 \\

        GLM-4.6v 
        & 0.66 / 0.31 / 0.38 & 0.70 / 0.27 / 0.34 & 0.70 / 0.21 / 0.30 
        & 0.86 / 0.03 / 0.47 & 0.86 / 0.03 / 0.41 & 0.88 / 0.03 / 0.35 \\

        \midrule
        \multicolumn{7}{c}{\textit{\textbf{Proprietary Models}}} \\
        \midrule
        
        Qwen-VL-Flash 
        & 0.58 / 0.40 / 0.49 & 0.63 / 0.35 / 0.43 & 0.65 / 0.27 / 0.37 
        & 0.76 / 0.09 / 0.63 & 0.82 / 0.07 / 0.57 & 0.83 / 0.06 / 0.54 \\
        
        Qwen-VL-Plus 
        & 0.59 / 0.41 / 0.47 & 0.63 / 0.35 / 0.43 & 0.65 / 0.28 / 0.38 
        & 0.77 / 0.08 / 0.63 & 0.79 / 0.08 / 0.58 & 0.84 / 0.06 / 0.56 \\

        GPT-5 Mini 
        & 0.82 / 0.04 / 0.23 & 0.82 / 0.04 / 0.24 & 0.86 / 0.01 / 0.16 
        & 0.97 / 0.04 / 0.30 & 0.97 / 0.03 / 0.29 & 0.98 / 0.03 / 0.27 \\
        
        GPT-5.1 
        & 0.74 / 0.09 / 0.22 & 0.73 / 0.08 / 0.23 & 0.80 / 0.03 / 0.15 
        & 0.94 / 0.06 / 0.32 & 0.96 / 0.03 / 0.24 & 0.96 / 0.03 / 0.25 \\

        Gemini 3 Flash 
        & 0.20 / 0.81 / 0.85 & 0.31 / 0.75 / 0.76 & 0.41 / 0.67 / 0.67 
        & 0.59 / 0.11 / 0.75 & 0.68 / 0.10 / 0.70 & 0.74 / 0.09 / 0.68 \\

        Gemini 3 Pro 
        & \textbf{0.16} / \textbf{0.87} / \textbf{0.90} & \textbf{0.25} / \textbf{0.79} / \textbf{0.82} & \textbf{0.35} / \textbf{0.70} / \textbf{0.73} 
        & \textbf{0.47} / \textbf{0.14} / \textbf{0.84} & \textbf{0.57} / \textbf{0.12} / \textbf{0.81} & \textbf{0.60} / \textbf{0.10} / \textbf{0.76} \\
        \bottomrule
    \end{tabular}%
    }
    \caption{Natural Language Reconstruction. Comparison on English and Chinese News. Format: NED ($\downarrow$) / BLEU ($\uparrow$) / ROUGE ($\uparrow$). Models are grouped by availability (Open-source vs. Proprietary).}
    \label{tab:news_results}
\end{table*}

\begin{table*}[t!]
    \centering
    \scriptsize
    \resizebox{0.8\textwidth}{!}{%
    \begin{tabular}{l|ccc|ccc|ccc}
        \toprule
        \textbf{Model} 
        & \multicolumn{3}{c|}{\textbf{C++}} 
        & \multicolumn{3}{c|}{\textbf{Java}} 
        & \multicolumn{3}{c}{\textbf{Python}} \\
        & $N=8$ & $N=12$ & $N=16$ & $N=8$ & $N=12$ & $N=16$ & $N=8$ & $N=12$ & $N=16$ \\
        \midrule
        \multicolumn{10}{c}{\textit{\textbf{Open-source Models}}} \\
        \midrule
        InternVL3.5-8B
        & .67 / .15 / .34 & .74 / .10 / .29 & .70 / .11 / .27 
        & .67 / .17 / .37 & .67 / .13 / .35 & .66 / .14 / .36 
        & .76 / .07 / .24 & .76 / .05 / .22 & .73 / .07 / .26 \\

        InternVL3.5-14B
        & .63 / .18 / .40 & .69 / .11 / .32 & .68 / .13 / .32 
        & .63 / .17 / .41 & .65 / .15 / .39 & .65 / .16 / .38 
        & .73 / .07 / .26 & .76 / .05 / .24 & .73 / .09 / .30 \\

        InternVL3.5-38B
        & .61 / .20 / .43 & .65 / .15 / .38 & .65 / .18 / .36 
        & .60 / .21 / .46 & .61 / .18 / .41 & .62 / .20 / .43 
        & .72 / .11 / .29 & .73 / .07 / .29 & .73 / .08 / .30 \\

        Mistral3-Reas-8B
        & .71 / .12 / .31 & .75 / .06 / .25 & .75 / .07 / .24 
        & .64 / .18 / .42 & .66 / .14 / .36 & .66 / .15 / .39 
        & .75 / .08 / .31 & .76 / .05 / .27 & .74 / .07 / .28 \\

        Mistral3-Reas-14B
        & .69 / .14 / .35 & .69 / .12 / .33 & .71 / .10 / .28 
        & .60 / .21 / .47 & .64 / .17 / .41 & .62 / .17 / .44 
        & .71 / .09 / .32 & .73 / .08 / .30 & .72 / .09 / .33 \\

        DeepSeek-OCR
        & .82 / .03 / .14 & .83 / .02 / .11 & .84 / .02 / .10 
        & .81 / .05 / .17 & .85 / .02 / .10 & .83 / .03 / .11 
        & .86 / .01 / .09 & .87 / .01 / .06 & .86 / .01 / .09 \\

        Hunyuan-OCR
        & .87 / .03 / .06 & .91 / .00 / .02 & .91 / .00 / .02 
        & .91 / .00 / .03 & .90 / .00 / .02 & .91 / .00 / .03 
        & .89 / .01 / .07 & .91 / .01 / .04 & .91 / .00 / .04 \\

        GLM-4.6v
        & .51 / .37 / .61 & .56 / .31 / .54 & .58 / .26 / .50 
        & .45 / .44 / .67 & .51 / .37 / .64 & .51 / .36 / .61 
        & .63 / .20 / .48 & .65 / .14 / .44 & .65 / .16 / .43 \\

        \midrule
        \multicolumn{10}{c}{\textit{\textbf{Proprietary Models}}} \\
        \midrule
        
        Qwen-VL-Flash
        & .47 / .43 / .66 & .48 / .37 / .62 & .54 / .31 / .54 
        & .42 / .48 / .74 & .48 / .45 / .68 & .55 / .41 / .63 
        & .57 / .32 / .55 & .56 / .25 / .55 & .60 / .22 / .51 \\

        Qwen-VL-Plus
        & .47 / .43 / .66 & .53 / .35 / .59 & .55 / .33 / .55 
        & .42 / .49 / .73 & .47 / .45 / .70 & .53 / .44 / .65 
        & .59 / .31 / .56 & .58 / .25 / .54 & .58 / .20 / .48 \\

        GPT-5 Mini
        & .74 / .13 / .30 & .74 / .09 / .25 & .78 / .08 / .21 
        & .75 / .15 / .38 & .79 / .08 / .29 & .74 / .13 / .36 
        & .98 / .04 / .19 & .80 / .04 / .23 & .75 / .08 / .28 \\
        
        GPT-5.1
        & .69 / .14 / .29 & .76 / .05 / .21 & .76 / .06 / .21 
        & .63 / .14 / .37 & .71 / .07 / .26 & .70 / .10 / .30 
        & .76 / .05 / .16 & .77 / .05 / .18 & .78 / .05 / .22 \\

        Gemini 3 Flash
        & .21 / .73 / .91 & .23 / .68 / .89 & .29 / .66 / .85 
        & .20 / .78 / .92 & .21 / .78 / .91 & .23 / .71 / .89 
        & .23 / .68 / .86 & .32 / .61 / .79 & .30 / .58 / .80 \\

        Gemini 3 Pro
        & \textbf{.20} / \textbf{.78} / \textbf{.92} & \textbf{.21} / \textbf{.76} / \textbf{.90} & \textbf{.25} / \textbf{.74} / \textbf{.88} 
        & \textbf{.18} / \textbf{.84} / \textbf{.93} & \textbf{.19} / \textbf{.81} / \textbf{.92} & \textbf{.22} / \textbf{.77} / \textbf{.90} 
        & \textbf{.20} / \textbf{.72} / \textbf{.88} & \textbf{.19} / \textbf{.70} / \textbf{.88} & \textbf{.25} / \textbf{.64} / \textbf{.85} \\
        \bottomrule
    \end{tabular}%
    }
    \vspace{0.2cm}
    \\ \footnotesize \textit{*Leading zeros (e.g., 0.74) are omitted in this table for space efficiency.}
    \caption{Source Code Reconstruction Breakdown. Detailed metrics for C++, Java, and Python. Format: NED ($\downarrow$), BLEU ($\uparrow$), and ROUGE ($\uparrow$). Open-source and Proprietary models are separated. }
    \label{tab:code_results}
\end{table*}

\begin{table}[t!]
    \centering
    \scriptsize
    \resizebox{\columnwidth}{!}{%
    \begin{tabular}{l|ccc}
        \toprule
        \multicolumn{4}{c}{\textbf{Category: Structured Table Data (Not Text/Code)}} \\ 
        \midrule
        \textbf{Model} 
        & \textbf{$N=8$} & \textbf{$N=12$} & \textbf{$N=16$} \\
        \midrule
        \multicolumn{4}{c}{\textit{\textbf{Open-source Models}}} \\
        \midrule
        InternVL3.5-8B
        & 0.85 / 0.06 / 0.10 & 0.83 / 0.07 / 0.09 & 0.84 / 0.03 / 0.09 \\
        
        InternVL3.5-14B
        & 0.82 / 0.03 / 0.12 & 0.82 / 0.04 / 0.11 & 0.84 / 0.02 / 0.09 \\
        
        InternVL3.5-38B
        & 0.80 / 0.06 / 0.12 & 0.79 / 0.05 / 0.12 & 0.79 / 0.05 / 0.11 \\

        Mistral3-Reas-8B
        & 0.82 / 0.05 / 0.20 & 0.84 / 0.03 / 0.16 & 0.83 / 0.05 / 0.19 \\

        Mistral3-Reas-14B
        & 0.83 / 0.03 / 0.19 & 0.84 / 0.05 / 0.18 & 0.84 / 0.03 / 0.16 \\

        DeepSeek-OCR
        & 0.87 / 0.01 / 0.07 & 0.85 / 0.00 / 0.06 & 0.86 / 0.01 / 0.06 \\

        Hunyuan-OCR
        & 0.80 / 0.09 / 0.30 & 0.81 / 0.04 / 0.30 & 0.83 / 0.03 / 0.22 \\

        GLM-4.6v
        & 0.75 / 0.04 / 0.23 & 0.79 / 0.04 / 0.19 & 0.82 / 0.05 / 0.16 \\

        \midrule
        \multicolumn{4}{c}{\textit{\textbf{Proprietary Models}}} \\
        \midrule
        
        Qwen-VL-Flash
        & 0.62 / 0.15 / 0.49 & 0.66 / 0.12 / 0.44 & 0.63 / 0.12 / 0.48 \\
        
        Qwen-VL-Plus
        & 0.61 / 0.17 / 0.51 & 0.68 / 0.12 / 0.44 & 0.67 / 0.10 / 0.43 \\

        GPT-5 Mini
        & 0.84 / 0.06 / 0.25 & 0.85 / 0.06 / 0.24 & 0.85 / 0.05 / 0.24 \\
        
        GPT-5.1
        & 0.80 / 0.10 / 0.30 & 0.84 / 0.06 / 0.22 & 0.87 / 0.04 / 0.18 \\

        Gemini 3 Flash
        & \textbf{0.49} / \textbf{0.23} / \textbf{0.69} & \textbf{0.49} / \textbf{0.22} / \textbf{0.68} & \textbf{0.49} / \textbf{0.22} / \textbf{0.68} \\

        Gemini 3 Pro
        & 0.59 / 0.20 / 0.63 & 0.58 / 0.22 / 0.63 & 0.61 / 0.19 / 0.57 \\
        \bottomrule
    \end{tabular}%
    }
    \caption{Structured Data Reconstruction. Evaluation on tabular data. Format: NED ($\downarrow$) / TEDS ($\uparrow$) / ROUGE ($\uparrow$).}
    \label{tab:table_results}
\end{table}

\subsection{Performance Analysis} 
In this section, we conduct a multi-dimensional analysis of reconstruction performance. Our evaluation is structured into four key aspects: (1) Natural Language, covering general prose; (2) Source Code, focusing on syntactic logic; (3) Structured Data, assessing tabular processing; and (4) Granularity Impact, analyzing performance degradation as fragment counts increase.
Table~\ref{tab:summary_results_split} summarizes the overall performance across all categories. Gemini 3 Pro demonstrates the strongest resilience, achieving the lowest NED (0.33) and highest ROUGE (0.83) scores at the 8-fragment level, consistently outperforming other proprietary and open-source models.

\paragraph{Natural Language (Table~\ref{tab:news_results}).} 
We observe a marked performance disparity between languages, with models consistently scoring lower on Chinese News compared to English. 
This divergence stems partially from the high information density of Chinese logograms: unlike Latin scripts where redundancy is distributed across multi-letter words, a physical tear through a single Chinese character often obliterates its semantic identity, creating a harder reconstruction task~\cite{lan2025mcbe}. 
Furthermore, this numerical gap is amplified by metric sensitivity. Since metrics like BLEU and ROUGE rely on exact n-gram matching, the lack of explicit delimiters in Chinese means that even minor reconstruction errors can disrupt word segmentation boundaries, disproportionately penalizing the scores compared to English.

\paragraph{Source Code (Table~\ref{tab:code_results}).}
While recent studies have shown that MLLMs can effectively understand code rendered as images with substantial token reduction~\cite{shi2026codeocr,shi2024between,zeng2026codesumm}, our results reveal a performance hierarchy driven by syntax. Averaged across all models and fragment settings ($N \in \{8, 12, 16\}$), explicitly structured languages like Java (Avg. NED 0.59) and C++ (0.62) outperform Python (0.68). We attribute this to syntactic redundancy: explicit delimiters (curly braces `\{ \}`, sem icolons) act as visual anchors for alignment. Conversely, Python's whitespace dependence proves challenging as shredding disrupts spatial layout. Lacking explicit closures, models struggle to infer indentation and maintain logical scope, resulting in higher structural error rates.

\begin{figure*}[t]
    \centering
    \includegraphics[width=0.85\textwidth]{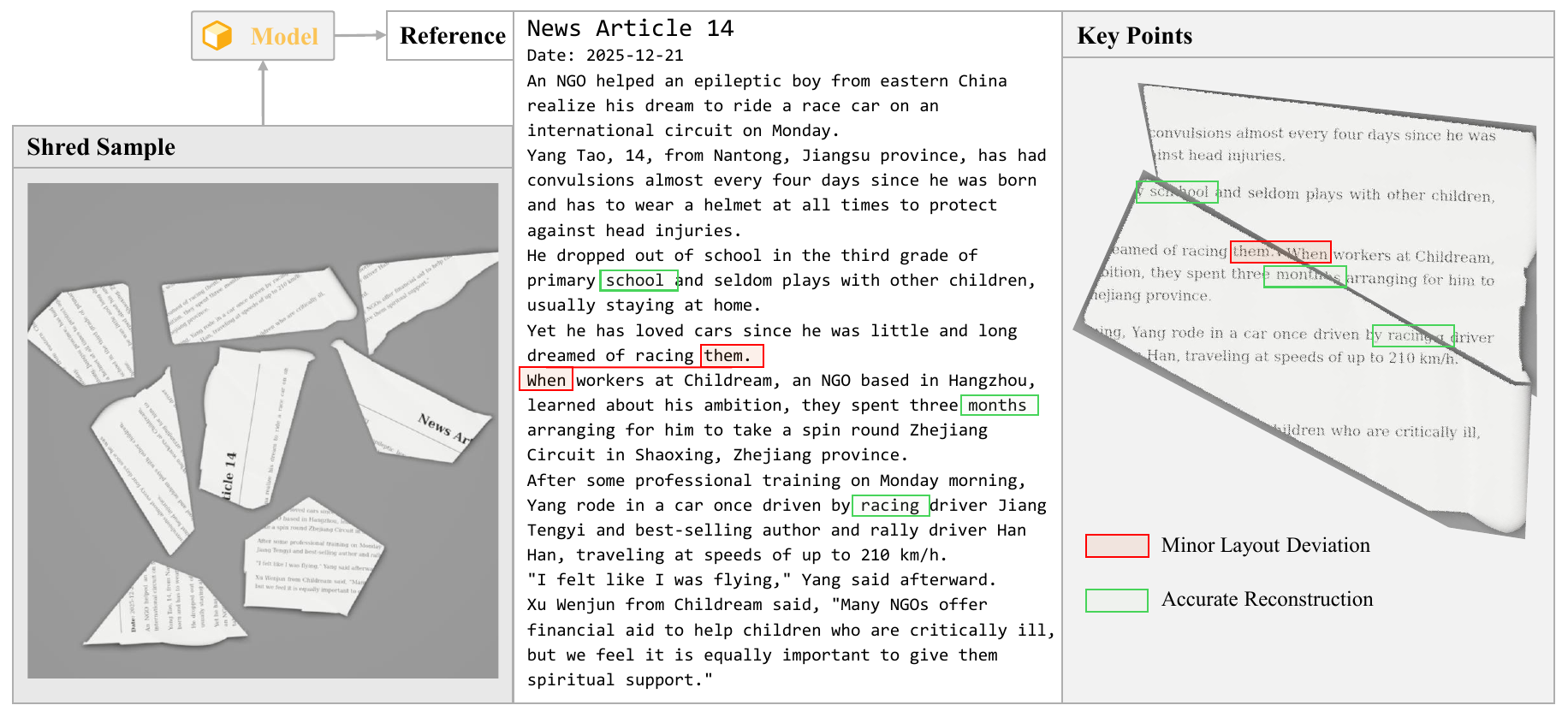}
    \caption{Good Case Study. The \textcolor{red}{red rectangle} highlights a minor layout inconsistency where the model interpreted a horizontal gap between fragments as a paragraph boundary (over-segmentation), despite the semantic continuity. The \textcolor{green}{green rectangle} demonstrates the model's robustness to physical fragmentation. Even though the characters are physically bisected, the model accurately synthesizes the disjointed visual cues to recover the complete word.}
    \label{fig:good_case_study}
\end{figure*}

\begin{figure*}[t]
    \centering
    \includegraphics[width=0.85\textwidth]{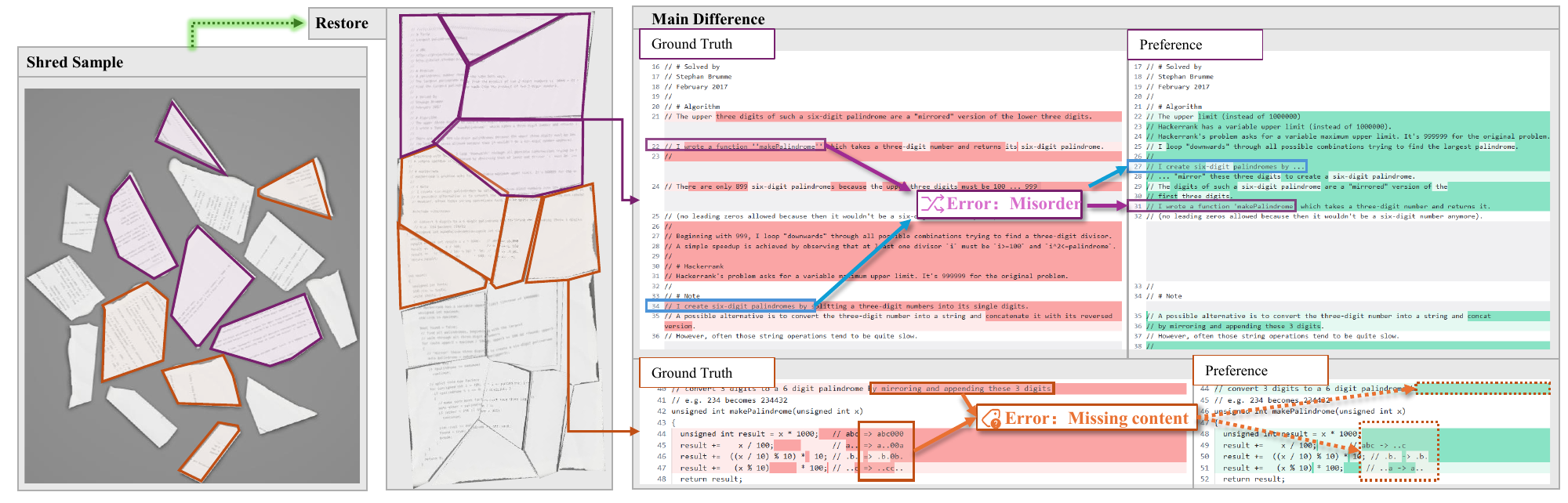}
    \caption{Bad Case Study. An example of code reconstruction failure. The \textcolor{magenta}{pink arrow} indicates an ordering error, where lines of code were structurally recognized but placed in the wrong logical sequence due to ambiguous visual cues. The \textcolor{orange}{orange box} highlights content loss, where a narrow strip containing code (e.g., \texttt{unsigned int}) was completely omitted, likely treated as visual noise.}
    \label{fig:bad_case_study}
\end{figure*}

\paragraph{Structured Data (Table~\ref{tab:table_results}).}
Table reconstruction presents a unique anomaly. While Gemini 3 Pro leads in text and code, Gemini 3 Flash significantly outperforms it on tabular data (NED 0.49 vs. 0.59). We suspect Flash's architecture might be more optimized for preserving rigid 2D spatial structures, whereas Pro prioritizes semantic flow, which can sometimes be detrimental when ``reading'' a non-linear table.

\subsection{Impact of Granularity}

We analyze performance as fragmentation increases ($N=8 \to 16$). As shown in Table~\ref{tab:summary_results_split}, performance generally degrades for most models. For instance, Qwen-VL-Plus sees an NED increase of 0.06 when moving from 8 to 16 fragments, while Gemini 3 Pro sees an increase of 0.08. Gemini 3 Pro nevertheless maintains the lowest overall NED across all three fragmentation levels.

\section{Qualitative Analysis}
\label{sec:discussion}

To understand the cognitive processes underlying reconstruction, we examine specific success and failure modes visualized in our case studies.

\subsection{Success Cases: Visual Semantic Bridging} 
Figure~\ref{fig:good_case_study} illustrates a successful reconstruction of a news article by Gemini 3 Pro. The model demonstrates two key capabilities. First, regarding visual closure (green highlights), the model successfully recovers words that are physically bisected by cuts. For example, the word ``school'' was split across two separate shards. The model did not merely OCR the fragments as ``sch'' and ``ool''; instead, it synthesized the disjointed visual cues to recover the complete token ``school''. This indicates the model is performing \emph{multimodal bridging}---using visual edge continuity to inform semantic prediction.  Second, regarding layout sensitivity (red highlight), the model is highly sensitive to physical gaps. In one instance, a horizontal gap between fragments was misinterpreted as a paragraph break (``When workers...''), leading to a minor layout deviation (over-segmentation) despite the text being semantically continuous.

\subsection{Failure Analysis: Where do MLLMs fail?}
Despite high aggregate scores, models struggle with global logic in complex documents, as seen in the code reconstruction example in Figure~\ref{fig:bad_case_study}.

Regarding ordering error (pink highlight), the most common error in code is \textit{logical misalignment}. The model correctly identified the text of lines 22 and 34 but swapped their order. Unlike prose, where semantic flow dictates order, code often consists of independent statements whose order is determined solely by algorithm logic, which is harder for the model to infer from visual shards alone. As for content loss (orange highlight), we observe instances of ``Hallucinated Deletion,'' where the model omits an entire line of code (e.g., line 43 `mirroring and appending this three digits`). This tends to happen with small, narrow strips of paper that contain only one line of text; the model may treat these isolated shards as visual noise or fail to integrate them into the larger context.
\vspace{-0.3cm}

\section{Conclusion}
\label{sec:conclusion}

In this work, we introduced \textsc{ShredBench}, a novel benchmark for evaluating the shredded content restoration capabilities of Multimodal LLMs. Our experiments across 756 documents and various modalities reveal that reconstruction is not merely a visual matching task but a complex reasoning challenge requiring the integration of visual cues (edge continuity) and semantic priors (language modeling). 

We find that Gemini 3 Pro establishes a new state-of-the-art, demonstrating superior resilience to fragmentation. However, significant challenges remain, particularly in strictly structured data (Tables), where even top models struggle to align disjointed cells. 

\section*{Limitations}

Our study operates under specific controlled constraints. First, our Voronoi-based cuts approximate irregular fragmentation but do not capture the full diversity of real-world tearing or cross-cut shredding mechanics. Second, although our pipeline simulates paper deformation in 3D, fragments are arranged without overlap, leaving occlusion and more complex folding outside the current evaluation setting. Third, regarding digital synthesis, while our ``ShredBench'' pipeline mimics physical fragmentation, domain shifts introduced by real-world environmental factors---such as variable lighting conditions and paper textures---remain an area for future exploration.

\section*{Acknowledgements}

We thank Haoran Gu for the helpful discussions. This paper is supported by the National Key Research and Development Program of China (Grant No. 2023YFB4503802) and the Natural Science Foundation of Shanghai (Grant No. 25ZR1401175).

% ...

\bibliography{custom}

\appendix

\section{Additional Evaluation: Metric Suitability for Code Restoration}
\label{sec:appendix_codebleu}

While standard string-matching metrics (such as NED, BLEU, and ROUGE) offer a robust general measure of text similarity, they may over-penalize benign formatting variations in strictly structured domains like source code. To provide a more structurally aware evaluation of model performance, we present additional experimental results utilizing CodeBLEU. Unlike standard n-gram metrics, CodeBLEU considers abstract syntax trees (AST) and semantic data flow, making it robust to whitespace and formatting differences that do not alter the underlying code logic.

Table \ref{tab:codebleu_appendix} presents the CodeBLEU scores of representative open-source and proprietary models on our full code dataset across C++, Java, and Python at varying fragmentation contexts ($N=8$, $N=12$, and $N=16$).

\begin{table*}[t] 
\centering
\small 
\caption{Source Code Restoration evaluated using CodeBLEU (Higher is better).}
\label{tab:codebleu_appendix}
\resizebox{0.9\textwidth}{!}{ 
\begin{tabular}{lccccccccc}
\toprule
\multirow{2}{*}{\textbf{Model}} & \multicolumn{3}{c}{\textbf{C++}} & \multicolumn{3}{c}{\textbf{Java}} & \multicolumn{3}{c}{\textbf{Python}} \\
\cmidrule(lr){2-4} \cmidrule(lr){5-7} \cmidrule(lr){8-10}
& $N=8$ & $N=12$ & $N=16$ & $N=8$ & $N=12$ & $N=16$ & $N=8$ & $N=12$ & $N=16$ \\
\midrule
\multicolumn{10}{l}{\textit{Open-source Models}} \\
InternVL3.5-8B & 0.32 & 0.28 & 0.25 & 0.34 & 0.31 & 0.31 & 0.22 & 0.20 & 0.25 \\
InternVL3.5-14B & 0.35 & 0.28 & 0.28 & 0.38 & 0.33 & 0.34 & 0.24 & 0.22 & 0.26 \\
InternVL3.5-38B & 0.38 & 0.32 & 0.34 & 0.39 & 0.37 & 0.38 & 0.28 & 0.24 & 0.26 \\
Mistral3-Reas-8B & 0.26 & 0.21 & 0.21 & 0.33 & 0.31 & 0.31 & 0.22 & 0.22 & 0.22 \\
Mistral3-Reas-14B & 0.26 & 0.21 & 0.21 & 0.33 & 0.31 & 0.31 & 0.22 & 0.22 & 0.22 \\
DeepSeek-OCR & 0.23 & 0.18 & 0.21 & 0.24 & 0.18 & 0.19 & 0.16 & 0.15 & 0.16 \\
Hunyuan-OCR & 0.12 & 0.12 & 0.11 & 0.11 & 0.07 & 0.09 & 0.10 & 0.08 & 0.06 \\
\midrule
\multicolumn{10}{l}{\textit{Proprietary Models}} \\
GPT-5 Mini & 0.26 & 0.22 & 0.19 & 0.34 & 0.27 & 0.31 & 0.17 & 0.19 & 0.22 \\
GPT-5.1 & 0.28 & 0.20 & 0.21 & 0.33 & 0.26 & 0.29 & 0.23 & 0.20 & 0.25 \\
Gemini 3 Flash & 0.79 & 0.76 & 0.73 & 0.86 & 0.84 & 0.81 & 0.74 & 0.68 & 0.68 \\
Gemini 3 Pro & 0.83 & 0.79 & 0.79 & 0.87 & 0.85 & 0.83 & 0.77 & 0.77 & 0.71 \\
\bottomrule
\end{tabular}
}
\end{table*}

\textbf{Discussion and Analysis:}

As demonstrated in Table \ref{tab:codebleu_appendix}, transitioning to an AST-aware metric highlights significant disparities in structural code restoration capabilities. The proprietary models, specifically Gemini 3 Pro and Gemini 3 Flash, exhibit exceptional structural fidelity, consistently achieving the highest scores across all languages and context lengths. This validates their robustness in complex structural formatting tasks over standard string-matching methods.

Among the open-source candidates, the InternVL3.5 series maintains strong baseline performance (peaking at $0.39$ on Java for the 38B model), effectively demonstrating positive

\section{Reproducibility and Evaluation Protocols}
\label{sec:appendix_reproducibility}

To ensure complete methodological transparency and facilitate future research, we detail the technical specifications of our evaluation pipeline and data generation process. We commit to open-sourcing our entire code repository—encompassing data generation, 3D rendering, and inference scripts—upon publication.

\subsection{Model Inference Protocol}
All evaluations were conducted using a consistent zero-shot system prompt. This prompt explicitly instructs the models to mentally ``stitch'' the fragments together and perform verbatim transcription, while strictly ignoring physical artifacts such as shadows, tears, and noise. 

To ensure deterministic and reproducible outputs across all evaluated model APIs, the decoding temperature was set to zero (or the minimum supported value). Furthermore, a rigorous post-processing script was applied to the raw model outputs to strip non-content artifacts (e.g., markdown tags, extraneous whitespace). This guarantees that our evaluation metrics (ROUGE, NED, TEDS, and CodeBLEU) exclusively reflect the accuracy of the restored document content.

\subsection{Data Generation and Physical Simulation}
For the visual inputs, we adopted a ``single composite image'' approach. The unordered document fragments were rendered onto a $4096 \times 4096$ high-resolution canvas using the Blender Cycles engine with global illumination to simulate realistic scanning environments. The original text documents were initially rendered at a width of 1600px with a 28px font size. The final composite images were subsequently resized to a maximum dimension of 2048px for model inference, striking a balance between preserving fine-grained visual perception and adhering to the models' visual token limits.

The physical complexity of the shredded documents is governed by the following simulation parameters:
\begin{itemize}[leftmargin=*]
    \item \textbf{Spatial Arrangement:} Fragments were subjected to random Z-axis rotations ranging from $0^\circ$ to $360^\circ$.
    \item \textbf{Irregular Boundaries:} Natural tearing edges were generated via Voronoi tessellation using $N \in \{8, 12, 16\}$ seed points.
    \item \textbf{Paper Deformation:} 3D physical artifacts were simulated using a Solidify modifier (thickness = $0.002$) combined with a two-tier displacement strategy. Large-scale paper waves were generated using a Marble texture (noise scale = $1.5$, strength = $0.15$), while sharp micro-crumples were applied using a Musgrave texture (noise scale = $8.0$, strength = $0.02$).
\end{itemize}

\begin{table*}[htbp]
\centering
\caption{Comparison of Reconstruction Performance on Real vs. Nonsense Text ($N=16$ Fragments). ``Real'' refers to the English News dataset, while ``Nonsense'' represents the randomized control text. $\Delta$ ROUGE denotes the absolute performance drop.}
\label{tab:ablation_results}
\resizebox{0.9\textwidth}{!}{
\begin{tabular}{l cc cc cc c}
\toprule
\multirow{2}{*}{\textbf{Model}} & \multicolumn{2}{c}{\textbf{NED} ($\downarrow$)} & \multicolumn{2}{c}{\textbf{BLEU} ($\uparrow$)} & \multicolumn{2}{c}{\textbf{ROUGE} ($\uparrow$)} & \multirow{2}{*}{$\Delta$ \textbf{ROUGE}} \\
\cmidrule(lr){2-3} \cmidrule(lr){4-5} \cmidrule(lr){6-7}
& Real & Nonsense & Real & Nonsense & Real & Nonsense & \\
\midrule
Gemini 3 Pro   & 0.35 & 0.65 & 0.70 & 0.39 & 0.73 & 0.33 & $-0.40$ \\
Gemini 3 Flash & 0.41 & 0.71 & 0.67 & 0.35 & 0.67 & 0.29 & $-0.38$ \\
Qwen-VL-Plus   & 0.65 & 0.75 & 0.28 & 0.06 & 0.38 & 0.13 & $-0.25$ \\
Qwen-VL-Flash  & 0.65 & 0.78 & 0.27 & 0.03 & 0.37 & 0.12 & $-0.25$ \\
GLM-4.6v       & 0.70 & 0.74 & 0.21 & 0.17 & 0.30 & 0.18 & $-0.12$ \\
GPT-5.1        & 0.80 & 0.81 & 0.03 & 0.00 & 0.15 & 0.08 & $-0.07$ \\
GPT-5 Mini     & 0.86 & 0.82 & 0.01 & 0.00 & 0.16 & 0.08 & $-0.08$ \\
\bottomrule
\end{tabular}
}
\end{table*}

\section{Ablation Study: Semantic Reasoning vs. Visual Matching}
\label{sec:appendix_ablation}

To determine whether models solve the fragmented document reconstruction task via semantic reasoning or by merely exploiting visual artifacts (e.g., edge matching), we conducted a controlled ablation experiment.

We generated a \textbf{Control Dataset} consisting of 50 documents using randomized ``nonsense'' text (e.g., ``the circumstances eligendi...''). We strictly preserved the exact layout, character length distribution, and font settings of the original English News dataset. The hardest fragmentation granularity ($N=16$) was applied using our physics-based pipeline. Our hypothesis is straightforward: if models rely primarily on visual edge matching (jigsaw solving), their performance on ``Nonsense'' text should be comparable to ``Real'' text. Conversely, if they depend on semantic language priors, their performance on ``Nonsense'' text should collapse due to the absence of semantic context needed to bridge visual discontinuities.

We evaluated seven representative models under this setting. As shown in Table \ref{tab:ablation_results}, performance dropped precipitously across all metrics when semantic meaning was removed. For instance, Gemini 3 Pro (the state-of-the-art model) achieves a high ROUGE score of 0.73 on real text, but this score collapses to 0.33 on nonsense text, accompanied by an NED degradation from 0.35 to 0.65. Gemini 3 Flash exhibits a similar decline ($\Delta$ ROUGE = $-0.38$).

Crucially, we observe a \textbf{Convergence of Failure}: on the nonsense dataset, all models degrade to a similarly low performance tier (NED ranging from 0.65 to 0.82). This indicates that without semantic cues, even the most capable models cannot effectively reconstruct the document based on visual features alone. The substantial performance gap confirms that visual artifacts are insufficient for reconstruction in \textsc{ShredBench}, success necessitates strong semantic reasoning.

\end{document}